\documentclass{article}

\usepackage{PRIMEarxiv}

\usepackage[utf8]{inputenc} %
\usepackage[T1]{fontenc}    %
\usepackage{hyperref}       %
\usepackage{url}            %
\usepackage{booktabs}       %
\usepackage{amsfonts}       %
\usepackage{nicefrac}       %
\usepackage{microtype}      %
\usepackage{lipsum}
\usepackage{fancyhdr}       %
\usepackage{graphicx}       %
\graphicspath{{media/}}     %
\usepackage{multirow}
\usepackage{caption}
\usepackage{subcaption}

\title{ReDDIT: Regret Detection and Domain Identification from Text}

\author{
  Fazlourrahman Balouchzahi\textsuperscript{\rm a}, Sabur Butt \textsuperscript{\rm b}, Grigori Sidorov\textsuperscript{\rm c}, and Alexander Gelbukh\textsuperscript{\rm d} \\
  Instituto Politécnico Nacional (IPN), Center for Computing Research (CIC), Mexico\\
  \texttt{\{\textsuperscript{\rm a}fbalouchzahi2021, \textsuperscript{\rm b}sbutt20211, \textsuperscript{\rm c}sidorov, \textsuperscript{\rm d}gelbukh\}@cic.ipn.mx} \\
}

\begin{document}
\maketitle

\begin{abstract}
In this paper, we present a study of regret and its expression on social media platforms. Specifically, we present a novel dataset of Reddit texts that have been classified into three classes: Regret by Action, Regret by Inaction, and No Regret. We then use this dataset to investigate the language used to express regret on Reddit and to identify the domains of text that are most commonly associated with regret. Our findings show that Reddit users are most likely to express regret for past actions, particularly in the domain of relationships. We also found that deep learning models using GloVe embedding outperformed other models in all experiments, indicating the effectiveness of GloVe for representing the meaning and context of words in the domain of regret. Overall, our study provides valuable insights into the nature and prevalence of regret on social media, as well as the potential of deep learning and word embeddings for analyzing and understanding emotional language in online text. These findings have implications for the development of natural language processing algorithms and the design of social media platforms that support emotional expression and communication.
\end{abstract}


\section{Introduction}~\label{Introduction}
Regret is a negative emotion that arises in response to events or circumstances that a person wishes had turned out differently~\cite{zeelenberg1998experience}. It is often associated with feelings of remorse, disappointment, and self-blame~\cite{budjanovcanin2022regretting, michenaud2008applying}. However, it can also push a person towards revised decision-making, leading to improved life circumstances~\cite{zeelenberg1999anticipated}. Regret is a common emotion that has been shown to have a significant impact on human behavior and decision-making, and is considered to be an essential emotion to explore and understand~\cite{zeelenberg2002inaction}.

Research on regret has largely focused on its role in decision-making, with studies showing that regret is a key factor in how people evaluate the outcomes of their decisions. Regret is typically associated with the concept of decision-making under uncertainty, with people not only caring about what they receive but also considering what they might have received if they had made a different decision~\cite{bleichrodt2010quantitative}. This can lead to feelings of regret when the outcome of a decision is less desirable than what was expected~\cite{diecidue2017regret}.

Regret is often triggered by actions or inactions that a person takes or fails to take in the past~\cite{gilovich1995experience, hattiangadi1995failing, roese2005we}. In some cases, regret may be more painful in the short term if it is the result of an action, but may last longer and be more problematic in the long term if it is the result of inaction~\cite{gilovich1995experience,hattiangadi1995failing}. Identifying the origin of regret is therefore important in understanding its impact on behavior and decision-making.

The influence of regret on behavior and decisions may also vary depending on the domain in which it is experienced. Research has shown that common domains for regret among Americans include education, career, romance, parenting, self-improvement, and leisure~\cite{roese2005we}.

Despite its importance and prevalence, regret has received little attention in the field of natural language processing. In this paper, we aim to address this gap by proposing a computational approach to studying regret. Our goal is to shed light on the role of regret in human emotions and decision-making, and provide a foundation for further research in this area. By doing so, we hope to better understand the ways in which regret impacts our lives and decision-making processes, and to develop more effective ways of dealing with it.

The main focus of this paper is to present and analyze the first-ever annotated dataset on regret detection and domain identification in English texts from Reddit. The dataset was created by scraping Reddit and primarily contains texts from three Subreddits: "regret", "regretfulparents", and "confession" in the timeline of 01-01-2000 to 10-09-2022. We modelled the task of regret detection and domain identification by first, categorizing Reddit posts into three classes, namely: "Action" (regret by action), "Inaction" (regret by in action), "No regret", and second, identifying the domain of the text (to see the annotation guidelines and sample refer to \ref{Annotation guideline}~Annotation guideline). Further, several baseline and state-of-the-art experiments were performed to benchmark the dataset.

Our research indicates that Reddit users are more likely to express regret about actions they have taken in the past, particularly in the context of relationships. We also found the effectiveness of GloVe (Global Vectors for Word Representation)~\cite{pennington2014glove} for representing the meaning and context of words in the domain of regret in such a way that deep learning models using GloVe word embeddings outperformed other models in all experiments.

The main contributions of this paper are:
\begin{itemize}
    \item A study of regret as an individual emotion from a computational perspective,
    \item The development of the first multi-class regret detection dataset,
    \item An analysis of the origins and domains of different types of regret,
    \item A range of machine learning and deep learning models to benchmark the dataset,
    \item A discussion of the performance of the learning models and a detailed error analysis.
\end{itemize}

\section{Related Work} \label{Related Work}
Regret is a common emotion in daily life~\cite{saffrey2008praise}, and according to a report on verbal expressions, it is the second most frequent emotion after love~\cite{shimanoff1984commonly}. Despite its frequency and significant impact on human life and behavior, regret has rarely been studied computationally.

In the few studies that have been done, regret has been explored in the context of social media analysis using natural language processing, in terms of identifying and analyzing regrettable posts. These studies aimed to automatically detect potentially regrettable disclosures, such as hate speech, profanity, offensive language, and private or sensitive information that could negatively impact a user's reputation and life~\cite{shulman2021poster,simo2022regrets}.

In a study by Wang et al.~\cite{wang2019donttweetthis}, the authors explored the phenomenon of users sometimes revealing too much information or unintentionally releasing regretful messages on social media. They proposed that these instances are more likely to occur when users are careless, emotional, or unaware of privacy risks. The authors collected a large number of tweets and labeled them into five categories: Very sensitive, Sensitive, Little Sensitive, Maybe, and Nonsensitive. They then performed a range of analyses on the tweets, the details of which can be found in the reference paper.

Caliskan et al.~\cite{caliskan2014privacy} proposed a method for identifying private content in text using a combination of topic modeling, named entity recognition, privacy ontology, sentiment analysis, and text normalization. They collected approximately 500,000 tweets from 100,000 users and modeled the task as a binary user profiling problem, where each user was identified as either a sensitive/private information spreader or not. The authors achieved an accuracy of 95.45\% in their best-performing experiment. They also annotated the tweets into three labels based on their privacy score, with higher numbers indicating a higher level of sensitivity. In this three-class classification task, they obtained an accuracy of 69.63\%.

The mentioned works indicate that regret has not been studied as an emotion in its own right. Instead, it has been examined from a privacy-related perspective, with a focus on understanding the spread of sensitive content on social media networks and the potential for regret among users.

On the other hand, previous research on emotion analysis has primarily treated it as a multilabel text classification task. However, regret has not been included in these studies, which have focused on emotions such as happiness, anger, disgust, fear, sadness, and surprise. A complete study of these literatures is presented by~\cite{ashraf2022multi}.

Some of these studies on emotion detection are summarized as follows: 
In the study by Halim et al.~\cite{halim2020machine}, a methodology for detecting emotions in email messages was proposed. The framework utilized autonomous learning techniques and incorporated three machine learning classifiers (ANN, SVM, and RF) along with three feature selection algorithms. The aim was to identify six emotional states (neutral, happy, sad, angry, positively surprised, and negatively surprised) in the text of email messages.

In~\cite{li2020interactive}, a model called IDS-ECM was introduced for predicting emotions in textual dialogue. The study compared textual dialogue emotion analysis to generic textual emotion analysis and identified context-dependence, contagion, and persistence as key features of the former.

Neural network-based models, such as bi-LSTM, CNN, and LSTM, have been shown to outperform feature-based supervised models like SVM and MaxEnt in emotion classification tasks~\cite{barnes2017assessing,schuff2017annotation}. In a recent study, Basiri et al.~\cite{basiri2021abcdm} proposed a CNN-RNN deep bidirectional model with attention mechanisms (ABCDM) to evaluate temporal information flow in both directions. ABCDM uses bidirectional LSTM and GRU layers to extract both past and future contexts, and attention mechanisms to focus on certain words. The use of convolution and pooling methods helps to minimize feature dimensionality and extract position-invariant local features. ABCDM showed state-of-the-art performance on sentiment polarity classification tasks using both long review and short tweet data, outperforming six previously suggested DNNs for sentiment analysis.

In a study conducted by Hassan et al.~\cite{hassan2021cross}, three approaches were examined for emotion classification in multilingual text: using intrinsically multilingual models, translating training data into the target language, and using a parallel corpus with automatic labeling. The research used English as the source language and Arabic and Spanish as the target languages. Various classification models, such as BERT and SVMs, were investigated and trained using different features. The results showed that BERT-based monolingual models trained on target language data outperformed the previous state-of-the-art by 4\% and 5\% in absolute Jaccard score for Arabic and Spanish, respectively. The BERT models achieved accuracies of 90\% and 80\% for Arabic and Spanish, respectively.

While regret is often associated with the past~\cite{gilovich1995experience}, it can also be related to the future~\cite{zeelenberg1999anticipated}. This is because regret often involves wishing that something had happened differently in the past~\cite{zeelenberg1998experience}, and this wish can be linked to how we expect things to unfold in the future. For example, if we regret not taking a certain opportunity in the past, we may fear that this will have negative consequences for us in the future. Similarly, if we regret something that we did in the past, we may worry that this will have negative repercussions for us in the future. In both of these cases, the emotion of regret is tied to our expectations for the future and can influence how we think and act.

In a similar vein, the emotion of hope can also be related to the future. Hope is an emotion that involves the anticipation of a good outcome or the possibility of something going well~\cite{balouchzahi2022polyhope}, while regret is a negative emotion that involves feeling sad or sorry about something that has already happened and cannot be changed. Both hope and regret can be experienced to varying degrees, and both can influence our thoughts and actions. Additionally, both hope and regret can be influenced by our experiences and our expectations for the future.

In a study on hope speech detection~\cite{balouchzahi2022polyhope}, researchers collected tweets and identified them as expressing hope or not. They then further classified the hope-related tweets into three categories: generalized, realistic, and unrealistic hopes. The authors experimented with several learning approaches for this task and found that a BERT transformer outperformed other methods, with averaged-macro F1-scores of 0.85 for the binary classification and 0.72 for the multiclass classification of hope.

\section{Dataset Development}\label{Dataset Development}
In this proposed work, we present a novel dataset for regret detection and domain identification in English posts from Reddit. We selected three subreddits - "regret", "regretfulparents", and "confession" - and scraped user posts from 1-1-2000 to 10-09-2022 using the Pushshift~\footnote{\url{https://github.com/pushshift/api}} API and the PMAW~\footnote{\url{https://pypi.org/project/pmaw/}} framework. During the scraping process, we discarded empty or deleted posts, resulting in a dataset of 1782, 1021, and 187870 posts from the "regret", "regretfulparents", and "confession" subreddits, respectively. We randomly selected a sample of 2000 posts from the "confession" subreddit and merged it with the posts from the other two subreddits, resulting in 4803 unlabeled posts.

We then filtered the texts to remove posts with less than 10 or more than 500 words and removed duplicate rows, resulting in 3440 posts. However, these are not the final statistics of the dataset - for more details, please see the~\ref{Dataset statistics}, the dataset statistics section of this article.

\subsection{Annotation guideline}\label{Annotation guideline}
We provided the annotators with instructions and examples of different types of regret and their domains, as discussed in Section~\ref{Introduction}. These examples are presented in Table~\ref{tab:samples}.

\begin{itemize}
    \item \textbf{Regret Detection:} The aim of this subtask is to classify texts into three categories: "regret by action" (Action), "regret by inaction" (Inaction), and "No regret".
        \begin{itemize}
        \item \textbf{Regret by Action (Regret of doing something):}\\Regret by action is regretting a decision or choice or an action one has done in the past. One may regret that he has done something because it gives him undesired consequences in either the short or long term.
        
        \item \textbf{Regret by Inaction (Regret of NOT doing something):}\\ Regret by inaction, on the contrary, is a regret caused by a lack of decision or failure in doing something. One may regret that he has not done something and because of failure in doing that action, he has received undesired consequences in the short or long term.
        \item \textbf{No Regret:}\\ Text does not convoy any type of regret.

    \end{itemize}
    \item \textbf{Domain Identification:} In this subtask, each text will be identified and classified into one of pre-defined domains: "Education", "Health", "Career and Finance", "Romance and Relationships", and "Other Domains".
        \begin{itemize}
        \item \textbf{Education:}\\ A text's domain is Education if it discusses topics and ideas that are commonly associated with the field of education. This could include things like teaching methods, educational policy, learning theories, academic research, and other subjects that are relevant to the field of education. %
        \item \textbf{Health:}\\ A text's domain is "Health" if it discusses topics related to health, wellness, and medicine. This could include text that discusses diseases, treatments, medications, health care providers, and medical research. Additionally, a text's domain may be considered "Health" if it discusses health and wellness from a broader perspective, including topics like nutrition, exercise, and mental health. Overall, the key to identifying a text's domain as "Health" is to look for text that discusses topics that are related to maintaining and improving physical, mental, and emotional well-being.
        \item \textbf{Career and Finance:}\\A text's domain can be identified as "Career and Finance" if it discusses topics related to managing one's professional life and financial resources. This might include information about finding a job, building a career, managing finances, investing, and saving money. Keywords and phrases that are commonly associated with this domain include "career", "job," "finance," "investment," "savings", and "income." Additionally, a text that discusses topics like salary negotiation, career advancement, budgeting, and financial planning would also be indicative of the "Career and Finance" domain.
        \item \textbf{Romance and Relationships:}\\A text's domain can be identified as "Romance and Relationships" if it discusses topics related to romantic love, interpersonal relationships, parents, and human emotions. This can include text that discusses dating, marriage, infidelity, breakups, and other aspects of romantic and interpersonal relationships. Additionally, a text in this domain may discuss psychological theories and research related to love, attachment, and interpersonal relationships. Words and phrases like "love," "romance," "relationship," "heartbreak," and "intimacy" can all be indicative of a text's domain as Romance and Relationships.
        \item \textbf{Other domains:} If the text is in any domain other than the above-mentioned domains, e.g, Politics, Technology, etc.
    \end{itemize}
\end{itemize}

\begin{table}[ht]
    \centering
    \caption{Sample texts from the dataset}
    \resizebox{\columnwidth}{!}{%
    \begin{tabular}{l|c|c}
        \bf Text  &\bf Regret Detection &\bf Domain Identification  \\ \hline
         \begin{tabular}[c]{@{}l@{}}Ok so I've been falling hard for this girl recently and decided that today is\\ the day, do I asked her if she likes me via text. She said I'm a great guy but\\ she's not interested right now. Ugh, I feel like I don't know what to do anymore\\ :( just had to post this somewhere. \end{tabular}&Action & Romance and Relationships \\
         
         \hline
        
         \begin{tabular}[c]{@{}l@{}} I've lost so much weight, and everyone keeps saying how great I look. The\\ medication also makes me mildly euphoric. I know it's wrong, but I tried going\\ off the meds before, and I went through a withdrawal and started gaining weight.\end{tabular}&No Regret & Health \\
         
         \hline
        
         \begin{tabular}[c]{@{}l@{}}"I did not really understand that credit cards are fine as long as you pay it all\\ off on time. So I never applied for one back then because I heard that they are\\ bad for young people. Big mistake. I am trying to decrease the regret/self hate\\ for this oversight" \end{tabular}& Inaction&Career and Finance \\
         
         \hline
        
         \begin{tabular}[c]{@{}l@{}} I've made some shitty mistakes, like some really shitty ones. Sometimes I\\ think I'm a terrible person, but that would indicate that guilt makes me not a\\ terrible person. But thinking that excuses my problems is something that my\\ mind can't comprehend. I really just struggle with my thoughts sometimes.\end{tabular} & Action&Other Domains \\
         
         \hline
        
         \begin{tabular}[c]{@{}l@{}} I cheated on ten geometry lessons and two tests in last year.  I feel awful and \\never want to cheat on anything again, ever.  That feels better.\end{tabular} & Action&Education \\
         
         \hline
        
         \begin{tabular}[c]{@{}l@{}} Yep that's pretty much it. If they're jewish/asian mix it's all over. I will want to\\ marry her on the spot.\end{tabular}&No Regret & Romance and Relationships \\

    \end{tabular}
    }
    
    \label{tab:samples}
\end{table}

\subsection{Annotation procedure}
In this study, we recruited three annotators with backgrounds in IT and computer science and high proficiency in English to perform the annotation process. We conducted several individual meetings with the annotators to discuss the annotation guidelines in detail. The dataset was divided into four batches, and after each batch was labeled, the labeled samples were reviewed by the authors of the paper. During the annotation process, the annotators were instructed to carefully read each post and first identify the appropriate label for the type of regret (or lack of regret) expressed in the text. They were then asked to label the text with the most suitable domain, or with the "Other domains" label if the text did not fit into any of the predefined categories.

\subsection{Inter-annotator agreement}
Inspired by~\cite{balouchzahi2022polyhope}, Fleiss' kappa was used to assess the reliability of agreement between annotators. A Fleiss' kappa of 0.78 for regret detection and 0.84 for domain identification illustrate the strength and reliability of the proposed dataset.

\subsection{Dataset statistics}\label{Dataset statistics}

\begin{figure*}[ht]
    \centering
    \includegraphics[width=0.8\textwidth]{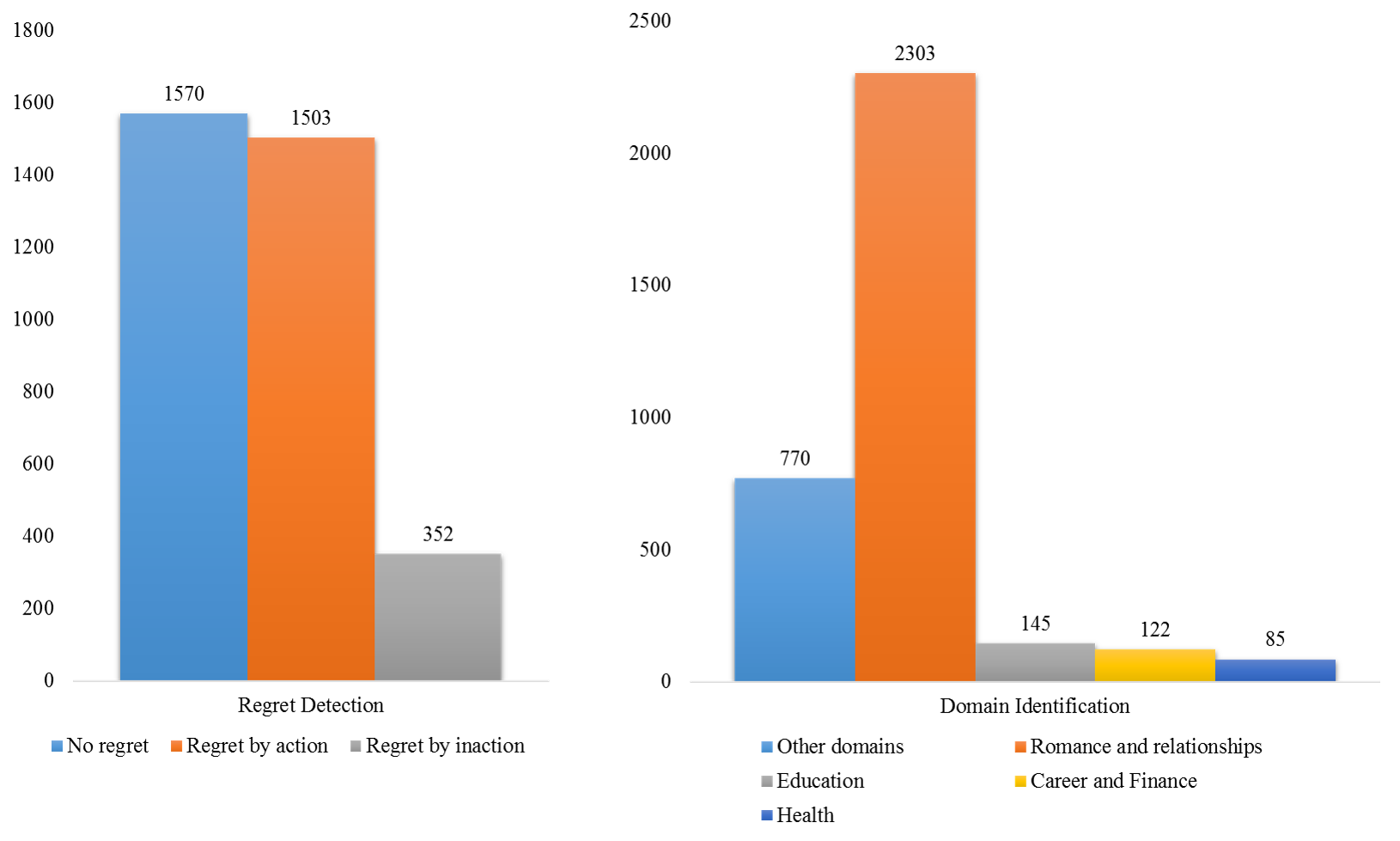}

    \caption{Label distribution for each subtask}
    \label{fig:labelDist}
\end{figure*}

We present the final statistics of our dataset in Table~\ref{tab:statistics} and a graphical representation in Figure~\ref{fig:labelDist}. Our analysis shows that Reddit users (specifically on the scraped subreddits) are more likely to express regret through actions rather than inaction. Additionally, the majority of posts in the dataset were related to romance and relationships. These statistics reveal an imbalanced distribution of labels for the regret detection subtask and a higher rate for the domain identification subtask. This imbalanced distribution of labels may impact the performance of learning models. A cross-subtask analysis of the data distribution, shown in Table~\ref{tab:crossSub}, reveals that users are less likely to express regret about decisions made regarding their health and other aspects of life, and are more likely to express regret about relationships. Furthermore, our analysis indicates that users are more likely to express regret about actions taken in relationships rather than inaction.

\begin{table}[ht]
\centering
\caption{Statistics of dataset}
\begin{tabular}{l|ll}

\textbf{Subtask}                                & \multicolumn{1}{l|}{\textbf{Classes}}          & \textbf{Count} \\ \hline
\multirow{3}{*}{\textbf{Regret Detection}}      & \multicolumn{1}{l|}{No Regret}                 & 1570           \\ 
                                                & \multicolumn{1}{l|}{Regret by Action}          & 1503           \\  
                                                & \multicolumn{1}{l|}{Regret by Inaction}        & 352            \\ \hline
\multirow{5}{*}{\textbf{Domain Identification}} & \multicolumn{1}{l|}{Other Domains}             & 770            \\ 
                                                & \multicolumn{1}{l|}{Romance and Relationships} & 2303           \\ 
                                                & \multicolumn{1}{l|}{Education}                 & 145            \\ 
                                                & \multicolumn{1}{l|}{Career and Finance}        & 122            \\  
                                                & \multicolumn{1}{l|}{Health}                    & 85             \\ \hline
\textbf{Total No. Posts}                        & \multicolumn{2}{c}{3425}                                       \\ 
\end{tabular}

\label{tab:statistics}

\end{table}

\begin{table}[ht]
\caption{Cross subtask distribution of labels}
\centering

\begin{tabular}{l|c|c|c}

\textbf{Classes}                   & \textbf{No Regret} & \textbf{Regret by   Action} & \textbf{Regret by   Inaction} \\ \hline
\textbf{Other Domains}             & 386                & 316                         & 68                            \\ 
\textbf{Romance and Relationships} & 1071               & 1007                        & 225                           \\ 
\textbf{Education}                 & 28                 & 85                          & 32                            \\ 
\textbf{Career and Finance}        & 38                 & 62                          & 22                            \\ 
\textbf{Health}                    & 47                 & 33                          & 5                             \\ 
\end{tabular}

\label{tab:crossSub}
\end{table}

\section{Benchmarks}
In this work, we used a set of traditional machine learning models as baselines, inspired by the work of Balouchzahi et al. (2022)~\cite{balouchzahi2022polyhope}. These baselines were used to evaluate the reliability of our dataset and annotations. In addition, we also evaluated two state-of-the-art deep learning models, namely, CNN and BiLSTM, to compare their performance with the traditional machine learning models. All the models were trained using five-fold cross-validation, and the average scores are reported in the~\ref{Results}. Results and Analysis section. Our main goal in this work was to introduce the task of regret detection using natural language processing techniques and to evaluate the feasibility of this task using our proposed dataset. Therefore, we only experimented with several simple baselines as well as state-of-the-art deep learning models, in order to validate our proposed approach and explore how neural network models can improve performance.

\subsection{Preprocessing}
Text preprocessing was performed by removing unicode characters, HTML/XML tags, URLs, and non-alphanumeric values. Contracted words were also expanded to their original forms (e.g. "I\'d" was expanded to "I would", and "could\'ve" was expanded to "could have"). Finally, stopwords were removed from the text.

\subsection{Traditional machine learning models}
Evaluating a new dataset with simple machine learning baselines is important as it provides a benchmark for comparison and helps to identify areas of improvement. It also helps to identify potential bias, outliers, and other issues in the data that can affect the performance of a machine learning model. Understanding the data, enable us later to build better models and improve the accuracy of the predictions. It also helps to compare the performance of different algorithms and find the best model for a given dataset.

Therefore, in this study, we evaluate our dataset using simple machine learning baselines to provide a benchmark for comparison and identify potential issues with the data. We used uni-grams TF-IDF vectors to train several traditional ML classifiers, including LR, SVM with Radial Basis Function (RBF) and linear kernels, RFC, XGB, and AdaBoost. All models were trained using default parameters to gain a better understanding of the performance of simple classifiers on the dataset. While we expect that further feature engineering and hyperparameter tuning could improve performance, these steps were not pursued in this study and are left for future work.

\subsection{Deep learning models}
CNN and BiLSTM models are both commonly used in the field of natural language processing for text classification tasks. CNNs are a type of neural network that is particularly well-suited for processing data with a spatial structure, such as images or text. In the context of text classification, CNNs can be used to automatically learn features from the input text, which can then be used to make predictions about the class of the text~\cite{lai2015recurrent}.

BiLSTMs, on the other hand, are a type of recurrent neural network that is capable of modeling sequential data, such as text. Unlike traditional LSTM networks, which only process the input sequence in one direction, BiLSTMs are able to consider the input sequence in both forward and backward directions, allowing them to capture contextual information from the entire sequence. This can be useful for text classification tasks, as it allows the model to consider the entire input text when making predictions~\cite{liu2019bidirectional}.

The main difference between CNNs and BiLSTMs for text classification is the type of information they are able to capture from the input data. CNNs are able to learn features automatically from the input text, while BiLSTMs are able to consider contextual information from the entire input sequence. 

Both CNN and BiLSTM models can be effective for text classification, and the comparison between the performance of these two models for the regret detection and domain identification task must be interesting.

For this purpose, two widely used fastText and GloVe embedding were used in turn to train CNN and BiLSTM models. Texts were converted to sequences using Keras Tokenizer and then all sequences were padded to the maximum length of sequence in the dataset. The obtained sequences along with the embedding matrix from word embedding as weights were employed to train DL models for 20 epochs for each fold. Table~\ref{tab:ParamDL} was borrowed from~\cite{balouchzahi2022polyhope} that indicates the parameters used in DL models.

\begin{table}[ht!]
\caption{\label{tab:ParamDL}Parameters for deep learning models}
\centering

\scalebox{0.9}{
\begin{tabular}{lcc}
 \bf Parameters & \bf CNN & \bf BiLSTM \\
\hline
    Epochs & 20 per fold & 20 per fold\\
    Optimizer & Adam & Adam\\
    loss & categorical crossentropy & categorical crossentropy\\
    filter sizes & [1,2,3,5] & -\\
    number of filters & 36 & -\\
    embedding size& 300 &300\\
    lr & 0.001 & 0.001\\
    Dropout & 0.1 &0.1\\
    Activation & softmax &softmax\\
    
\end{tabular}
              }

\end{table}

\section{Results and Analysis}\label{Results}
In this study, we used averaged weighted and macro scores to evaluate the performance of our learning models in a five-fold cross-validation manner. Since Reddit posts consist of both a title and main text, we trained each model first using only the main text, and then again using both the text and title, in order to compare the contribution of the title to the classification task.

\subsection{Regret Detection}
The performance of traditional machine learning baselines for the task of regret detection using only word uni-grams is presented in Table~\ref{tab:ResRegML}. The results using text+title show that adding more features from the titles did not improve the performance of these models, and generally only increased the dimensionality of the features. This lack of improvement may be due to the absence of any mechanism for feature engineering and hyperparameter tuning in the baselines. LR with an averaged-macro F1-score of 0.559 using only text outperformed other ML baselines for the regret detection task.

Table~\ref{tab:ResRegDL} reports the results obtained using deep learning models for the task of regret detection. Unlike the machine learning baselines, the results show that utilizing the title of posts along with the main text significantly improved the performance of the deep learning models. Additionally, the results show that in all experiments using deep learning models, both the CNN and BiLSTM models scored higher using the Glove embedding. The possible reasons for this are discussed in Section~\ref{Analysis}. Analysis. Eventually, CNN model with an averaged-macro F1-score of 0.715 obtained the state-of-the-art results for the task of multiclass regret detection using GloVe embedding and a combination of main text and title from Reddit posts.

\begin{table*}[ht]
\caption{\label{tab:ResRegML}Results for regret detection using traditional machine learning classifiers}
\centering

\begin{tabular}{l|ccc|ccc|c}
 \multirow{2}{*}{\bf Model} & \multicolumn{3}{c}{\bf Avg. Weighted Scores} & \multicolumn{3}{c}{\bf Avg. Macro Scores} & \multirow{2}{*}{\bf Accuracy} \\
        &Precision&Recall&F1-score&Precision&Recall&F1-score& \\

\hline
\multicolumn{8}{c}{\bf Regret Detection (only text)}\\
\hline
\bf LR &\bf 0.640 &\bf 0.648 &\bf 0.643 &\bf 0.571 &\bf 0.553 &\bf 0.559 &\bf 0.648 \\
SVM (rbf) &0.596 &0.664 &0.628 &0.442 &0.493 &0.466 &0.664  \\
SVM (linear) &0.598 &0.600 &0.598 &0.518 &0.517 &0.517 &0.600 \\
RFC &0.595 &0.654 &0.620 &0.441 &0.487 &0.460 &0.654   \\
AdaBoost &0.598 &0.617 &0.601 &0.525 &0.494 &0.496 &0.617   \\
XGB &0.659 &0.655 &0.622 &0.667 &0.496 &0.483 &0.655   \\

\hline
\multicolumn{8}{c}{\bf Regret Detection (text + title)}\\
\hline
LR &0.622 &0.630 &0.625 &0.560 &0.538 &0.545 &0.630 \\
SVM (rbf) &0.582 &0.648 &0.613 &0.432 &0.481 &0.455 &0.648  \\
SVM (linear) &0.587 &0.590 &0.588 &0.519 &0.515 &0.516 &0.590 \\
RFC &0.578 &0.638 &0.603 &0.429 &0.474 &0.448 &0.637   \\
AdaBoost &0.576 &0.590 &0.576 &0.525 &0.480 &0.487 &0.590   \\
XGB &0.635 &0.622 &0.593 &0.664 &0.473 &0.462 &0.622   \\

\end{tabular}

\end{table*}
\begin{table*}[ht]
\caption{\label{tab:ResRegDL}Results for regret detection using deep learning models}
\centering
\resizebox{0.9\columnwidth}{!}{%
\begin{tabular}{l|c|ccc|ccc|c}
 \multirow{2}{*}{\bf Word Embedding}&\multirow{2}{*}{\bf Model} & \multicolumn{3}{c}{\bf Avg. Weighted Scores} & \multicolumn{3}{c}{\bf Avg. Macro Scores} & \multirow{2}{*}{\bf Accuracy} \\
                                 &&Precision&Recall&F1-score&Precision&Recall&F1-score& \\

\hline
\multicolumn{9}{c}{\bf Regret Detection (only text)}\\
\hline
\bf \multirow{2}{*}{GloVe}    & CNN &0.707 &0.716 &0.705 &0.664 &0.597 &0.612 &0.715  \\
\bf                              &BiLSTM &0.701 &0.709 &0.697 &0.653 &0.584 &0.596 &0.709  \\
\cline{2-9}
\bf \multirow{2}{*}{fastText}       & CNN &0.631 &0.641 &0.630 &0.569 &0.526 &0.533 &0.641  \\
\bf                              &BiLSTM &0.631 &0.641 &0.630 &0.560 &0.523 &0.529 &0.641  \\

\hline
\multicolumn{9}{c}{\bf Regret Detection (text + title)}\\
\hline
\bf \multirow{2}{*}{GloVe}   & \bf CNN &\bf 0.764 &\bf 0.763 &\bf 0.759 &\bf 0.759 &\bf 0.694 &\bf 0.715 &\bf 0.763  \\
\bf                              &BiLSTM &0.744 &0.744 &0.714 &0.713 &0.675 &0.688 &0.744  \\
\cline{2-9}
\bf \multirow{2}{*}{fastText}       & CNN &0.698 &0.698 &0.692 &0.688 &0.615 &0.636 &0.698  \\
\bf                              &BiLSTM &0.703 &0.696 &0.691 &0.686 &0.623 &0.640 &0.696  \\\

\end{tabular}
}

\end{table*}

\subsection{Domain Identification}
Tables~\ref{tab:ResDIML} and~\ref{tab:ResDIDL} present the performance of the machine learning baselines and deep learning models for the task of domain identification. Unlike to the regret detection subtask, the results show that the inclusion of the title slightly improved the performance of the machine learning models. LR classifier on text+title obtained an averaged-macro F1-score of 0.579 as the best-performing ML classifier for the task of domain identification. Furthermore, the results show that the deep learning models using the GloVe embedding again outperformed those using the fastText embedding. In the best performing result, BiLSTM with GloVe embedding outperformed all models for the task of domain identification with an averaged-macro F1 score of 0.629 on the combination of text from the title and main posts.

Additionally, the results show that all models had better performance on the regret detection task, which may be due to the smaller number of classes and a lower rate of imbalanced label distribution in the dataset.

\begin{table*}[ht!]
\caption{\label{tab:ResDIML}Results for domain identification using traditional machine learning classifiers}
\centering

\begin{tabular}{l|ccc|ccc|c}
 \multirow{2}{*}{\bf Model} & \multicolumn{3}{c}{\bf Avg. Weighted Scores} & \multicolumn{3}{c}{\bf Avg. Macro Scores} & \multirow{2}{*}{\bf Accuracy} \\
        &Precision&Recall&F1-score&Precision&Recall&F1-score& \\

\hline
\multicolumn{8}{c}{\bf Domain Identification (text)}\\
\hline
LR & 0.772 & 0.779 &0.769 & 0.666 & 0.509 & 0.552 & 0.779 \\
SVM (rbf) &0.688 &0.714 &0.638 &0.529 &0.270 &0.283 &0.714  \\
SVM (linear) &0.753 &0.756 &0.752 &0.623 &0.539 &0.564 &0.756 \\
RFC &0.720 &0.741 &0.687 &0.600 &0.300 &0.321 &0.741   \\
AdaBoost &0.556 &0.660 &0.574 &0.391 &0.327 &0.319 &0.660   \\
XGB &0.748 &0.767 &0.739 &0.664 &0.411 &0.466 &0.767   \\

\hline
\multicolumn{8}{c}{\bf Domain Identification (text + title)}\\
\hline
\bf LR &\bf 0.780 &\bf 0.782 &\bf 0.774 &\bf 0.712 &\bf 0.527 &\bf 0.579 &\bf 0.782 \\
SVM (rbf) &0.711 &0.737 &0.678 &0.545 &0.300 &0.325 &0.737  \\
SVM (linear) &0.758 &0.761 &0.757&0.629 &0.550 &0.572 &0.761 \\
RFC &0.714 &0.739 &0.688 &0.585 &0.307 &0.333 &0.739   \\
AdaBoost &0.558 &0.668 &0.582 &0.401 &0.352 &0.342 &0.668   \\
XGB &0.747 &0.768 &0.741 &0.653 &0.420 &0.474 &0.768   \\

\end{tabular}

\end{table*}
\begin{table*}[ht!]
\caption{\label{tab:ResDIDL}Results for domain identification using deep learning models}
\centering
\resizebox{0.9\columnwidth}{!}{%
\begin{tabular}{l|c|ccc|ccc|c}
 \multirow{2}{*}{\bf Word Embedding}&\multirow{2}{*}{\bf Model} & \multicolumn{3}{c}{\bf Avg. Weighted Scores} & \multicolumn{3}{c}{\bf Avg. Macro Scores} & \multirow{2}{*}{\bf Accuracy} \\
                                 &&Precision&Recall&F1-score&Precision&Recall&F1-score& \\

\hline
\multicolumn{9}{c}{\bf Domain Identification (only text)}\\
\hline
\bf \multirow{2}{*}{GloVe}    & CNN &0.806 &0.818 &0.806 &0.665 &0.551 &0.581 &0.818  \\
\bf                              &BiLSTM &0.796 &0.806 &0.797 &0.633 &0.548 &0.577 &0.806  \\
\cline{2-9}
\bf \multirow{2}{*}{fastText}       & CNN &0.710 &0.734 &0.718 &0.490 &0.398 &0.424 &0.735  \\
\bf                              &BiLSTM &0.745 &0.749 &0.740 &0.519 &0.435 &0.445 &0.749  \\

\hline
\multicolumn{9}{c}{\bf Domain Identification (text + title)}\\
\hline
\bf \multirow{2}{*}{GloVe}   & CNN &0.819 &0.828 &0.820 &0.687 &0.586 &0.619 &0.828  \\
\bf                              &\bf BiLSTM &\bf 0.814 &\bf 0.823 &\bf 0.816 &\bf 0.702 &\bf 0.588 &\bf 0.629 &\bf 0.823  \\
\cline{2-9}
\bf \multirow{2}{*}{fastText}       & CNN &0.721 &0.748 &0.729 &0.496 &0.405 &0.432 &0.748  \\
\bf                              &BiLSTM &0.775 &0.772 &0.770 &0.511 &0.481 &0.486 &0.772  \\\

\end{tabular}
}

\end{table*}
\subsection{Analysis}\label{Analysis}
Table~\ref{tab:bestPerforming} present the comparison between best-performing models for ML and DL models. The results show a huge improvement in performance using deep learning models.
Furthermore, based on observations, deep learning models using word embeddings often outperformed traditional machine learning classifiers with uni-grams that could be for several reasons, and some of these reasons are discussed below:

First, word embeddings provide a more detailed and accurate representation of the meaning and context of words compared to uni-grams, which are simply individual words treated as discrete tokens. Because word embeddings capture the relationship between words and their meanings, they can provide deep learning models with a more nuanced understanding of the text data they are trained on. This can help the models make more accurate predictions and classifications.

Second, deep learning models are able to learn complex patterns and relationships in the data, which allows them to outperform traditional machine learning algorithms that rely on more simplistic and rigid approaches. By using word embeddings as input, deep learning models are able to learn and capture the rich and complex patterns in the text data, which can improve their performance compared to traditional machine learning classifiers that rely on uni-grams.

Third, deep learning models can be trained on large amounts of data, which allows them to learn more robust and generalizable patterns in the data. This can improve their performance compared to traditional machine learning classifiers that we saw that in all cases adding title to the main texts improved the results for deep learning models while it had negative effects on machine learning baselines.

Hence, the combination of detailed word representations, complex pattern learning, and large-scale training data makes deep learning models using word embeddings more suitable models for the most of NLP tasks, and it was proven specifically for the regret detection and domain identification tasks as well.

In addition to this, the comparison of performance between all experiments reveal that deep learning models and on top of them models utilizing GloVe embedding outperformed the rest of learning models.

There are a few reasons why deep learning models had better performance using GloVe embeddings than fastText.

First, GloVe is a pre-trained word embedding model that is based on a large corpus of text data and trained using a specific algorithm called "co-occurrence matrix factorization"~\cite{pennington2014glove}. This means that it has already learned to map words to numerical vectors in a way that captures the meaning and context of the words in the corpus. As a result, GloVe embeddings can provide a more detailed and accurate representation of words and their meanings compared to fastText, which is a simpler model that uses a "bag of words" approach to represent text.

Second, deep learning models often require a large amount of data to train on and achieve good performance. Since GloVe is based on a very large corpus of text data, it can provide deep learning models with a rich and diverse set of pre-trained word vectors that can help improve their performance. In contrast, fastText is based on a much smaller corpus of text data, which may not be sufficient to support the training of deep learning models.

Overall, observation reveals that GloVe's pre-trained, fine-tunable, and detailed word vectors make it a better choice compared to fastText in the ReDDIT dataset.

\begin{table*}[ht]
\caption{Best performing models in each learning approach}
    \centering
    \resizebox{0.9\columnwidth}{!}{%
    \begin{tabular}{l|c|lcc}
         \bf Model&\bf Learning approach&\bf Training Data& \bf Averaged-weighted F1 & \bf Averaged-macro F1\\
         \hline
          \multicolumn{5}{c}{\bf Regret Detection}\\
        \hline
        LR& Machine learning&only text& 0.643 & 0.559\\
        CNN with GloVe& Deep learning& text + title & 0.759 & 0.715\\

         \hline
        \multicolumn{5}{c}{\bf Domain Identification}\\
        \hline
         LR& Machine learning&text + title& 0.774 & 0.579\\
         BiLSTM with GloVe& Deep learning &text + title& 0.816& 0.629\\

    \end{tabular}
    }
    
    \label{tab:bestPerforming}
\end{table*}

\section{Error Analysis}
Classifying regret as regret by action and regret by inaction can be difficult because these two types of regret are often expressed in similar ways and can be difficult to distinguish based on the words and phrases used. For example, someone may express regret about an action they took by saying "I wish I hadn't done that" or "I regret what I did," which could also be used to express regret about something they failed to do.

Additionally, the context in which regret is expressed can play a role in determining whether it is regret by action or regret by inaction. For example, a statement like "I regret not going to the party" could be interpreted as regret by inaction if the speaker is expressing regret about not attending the party, but it could be interpreted as regret by action if the speaker is expressing regret about not inviting someone else to the party.

Overall, the difficulties in classifying regret as regret by action and regret by inaction stem from the fact that these two types of regret can be expressed in similar ways and can be context-dependent. This can make it challenging to accurately identify and classify regret in text data.

For the error analyses, we combined the errors deducted in all k-folds of our best-performing model and analyzed the common patterns in them. 

\subsection{Errors of no-regret class}

The text originally labelled as no-regret, was misunderstood by our model as regret by inaction in the following scenarios:

\begin{itemize}
    \item Opinion: While a person is expressing his opinion, in actuality, he is not regretting, but expects the third person to regret it because of his/her inaction. This can also be understood as a future prediction about an individual or a group of people. 
    \item Fear: These texts contain the element of fear i.e. fear of missing out because of inaction. The person does not regret it at the moment but thinks the lost opportunity will cause regret. 
    \item Indifference: A person is indifferent about something that should have caused him to regret (for an action he did not commit), but he feels nothing about it. 
    \item Inner-Struggle: A person is describing his inner struggle of not-taking an action, in a certain situation but does not feel regret. 
\end{itemize}

The text originally labelled as no-regret, was misunderstood by our model as regret by action in the following scenarios:

\begin{itemize}
    \item Pleasure: The person engages in a situation that is wrong, unethical or contains a negative connotation to it, but instead of regretting, takes pleasure in it. 
    \item Negative emotion overlap: The person expresses negative emotions i.e. distrust, betrayal, fear, confusion, sadness or anger and is confused as regret by action. These texts identify actions but don't specifically highlight the feeling of regret. This scenario was the most encountered error for the no-regret class. 
    \item Indifference: A person is indifferent about something that should have caused him to regret (for an action he committed), but he feels nothing about it. 
\end{itemize}

\subsection{Errors of action class}
The text was annotated as regret by action but our model identified it as no regret. 

\begin{itemize} 
    \item Negative emotion overlap: The person expresses negative emotions i.e. embarrassment, fear, confusion, sadness or anger and is confused as no-regret. These texts identify actions and highlight the feeling of regret.
    \item Indifference: A person regrets an aspect of something he is done but is also indifferent about a part of his action. This causes the model to wrongly identify it as no regret. 
    
\end{itemize}

The text was annotated as regret by action but our model identified it as regret by inaction. 
\begin{itemize}
    \item Confusion between actions: While explaining the actions that the people committed in their respective past scenarios, they could not take the correct decisions and make proper actions to resolve the situation timely. Such cases, through representing regret by action, are confused with regret by inaction, because of several inaction words in the passage. It was noted that these passages can represent regret by action and inaction simultaneously. Regretting the wrong actions and regretting not making the correct actions. 
\end{itemize}

\subsection{Errors of inaction class}
The text was annotated as regret by inaction but our model identified it as no regret. 

\begin{itemize}
    \item Victim complex: Often the text suggests expressions of people being victims of their life events. They avoid taking action because they would rather become victims by avoiding responsibility.  
    \item Negative emotion overlap: The person expresses negative emotions i.e. embarrassment, fear, confusion, sadness or anger and is confused as no-regret. These texts don't identify actions and highlight the feeling of regret.  
    \item Procrastination: Spending time delaying taking an action and eventually ending up losing all time. Texts like these show no action and were sometimes confused with no regret. 
    \item Overlook: The text explains that the author failed to notice something and hence could not act. 
\end{itemize}

The text was annotated as regret by inaction but our model identified it as regret by action. 
\begin{itemize}
    \item Confusion between actions: While explaining the actions that the people committed in their respective past scenarios, they could not take the correct decisions and make proper actions to resolve the situation timely. Such cases, through representing regret by inaction, are confused by regret by action, because of several action words in the passage. It was noted that these passages can represent regret by action and inaction simultaneously. Regretting the wrong actions and regretting not making the correct actions. 
\end{itemize}

\section{Conclusion and Future Work}
Regret is an important and frequent emotion that can have a significant impact on an individual's well-being and mental health. By studying the ways in which people express regret on social media, researchers can learn more about the factors that contribute to regret, how it affects people, and how it can be managed or alleviated. This information can be used to develop interventions and support strategies for individuals who experience regret. In this view, this paper presents a description of a novel dataset for the task of regret detection and domain identification from Reddit posts. We have detailed our annotation guidelines and dataset development process. The statistics of the dataset showed that users on social media are more prone to share regret about their past relationships and actions. 

We also experimented with several machine learning baselines and state-of-the-art deep learning models to benchmark the dataset. In summary, deep learning models using word embeddings often outperformed traditional machine learning classifiers with uni-grams because they provide a more detailed and accurate representation of the meaning and context of words, and they were able to learn complex patterns and relationships in the data.

In future works, we would like to explore the different transformer-based models to see how a language model with higher context utilization compared to deep learning models can influence state-of-the-art results. Another possible direction for further analysis could be to investigate methods for addressing the imbalanced distribution of labels in the dataset. This could involve using techniques such as oversampling or undersampling to balance the distribution of labels, or using algorithms specifically designed for imbalanced datasets. Additionally, further analysis of the relationship between regret and specific domains, such as health and relationships, could provide valuable insights into the nature of regret and how it is expressed by social media users.

\section*{Acknowledgments}

The work was done with partial support from the Mexican Government through the grant A1-S-47854 of CONACYT, Mexico, grants 20220852 and 20220859 of the Secretaría de Investigación y Posgrado of the Instituto Politécnico Nacional, Mexico. The authors thank the CONACYT for the computing resources brought to them through the Plataforma de Aprendizaje Profundo para Tecnologías del Lenguaje of the Laboratorio de Supercómputo of the INAOE, Mexico and acknowledge the support of Microsoft through the Microsoft Latin America PhD Award.

\bibliographystyle{unsrt}  
\bibliography{references}

\end{document}